\documentclass[sigconf]{acmart}
\renewcommand\footnotetextcopyrightpermission[1]{} 
\usepackage{graphics}
\usepackage{adjustbox}
\usepackage{listings}
\usepackage{multirow}
\usepackage{varioref}
\usepackage[utf8]{inputenc}
\usepackage{xcolor}
\definecolor{codegreen}{rgb}{0,0.6,0}
\definecolor{codegray}{rgb}{0.5,0.5,0.5}
\definecolor{codepurple}{rgb}{0.58,0,0.82}
\definecolor{backcolour}{rgb}{0.95,0.95,0.92}
\lstdefinestyle{mystyle}{
    backgroundcolor=\color{backcolour},   
    commentstyle=\color{codegreen},
    keywordstyle=\color{magenta},
    numberstyle=\tiny\color{codegray},
    stringstyle=\color{codepurple},
    basicstyle=\ttfamily\footnotesize,
    breakatwhitespace=false,         
    breaklines=true,                 
    captionpos=b,                    
    keepspaces=true,                 
    numbers=left,                    
    numbersep=5pt,                  
    showspaces=false,                
    showstringspaces=false,
    showtabs=false,                  
    tabsize=2
}
\usepackage{enumerate}
\usepackage{enumitem}
\usepackage[caption = false]{subfig}
\usepackage{graphicx}
\usepackage{multicol, blindtext}
\usepackage{tabularx}

\newcommand{\m}{\textsf{MathBERT}}
\newcolumntype{Y}{>{\centering\arraybackslash}X}

\AtBeginDocument{%
  \providecommand\BibTeX{{%
    \normalfont B\kern-0.5em{\scshape i\kern-0.25em b}\kern-0.8em\TeX}}}

\begin{document}

\title{{\m}: A Pre-trained Language Model for General NLP Tasks in Mathematics Education}

\author{Jia Tracy Shen}
\affiliation{%
\institution{Penn State University}
\country{USA}
}
\email{jia.t.shen@gmail.com}

\author{Michiharu Yamashita}
\affiliation{%
\institution{Penn State University}
 \country{USA}
  }
\email{michiharu@psu.edu}

\author{Ethan Prihar}
\affiliation{%
  \institution{Worcester Polytechnic Institute}
 \country{USA}
}
\email{ebprihar@wpi.edu}

\author{Neil Heffernan}
\affiliation{%
  \institution{ASSISTments.org}
 \country{USA}
}
\email{neil@ASSISTments.org}

\author{Xintao Wu}
\affiliation{%
  \institution{University of Arkansas}
 \country{USA}
}
\email{xintaowu@uark.edu}

\author{Ben Graff}
\affiliation{%
  \institution{Stride, Inc}
 \country{USA}
}
\email{bgraff@k12.com}

\author{Dongwon Lee}
\affiliation{%
\institution{Penn State University}
 \country{USA}
  }
  \email{dongwon@psu.edu}

\renewcommand{\shortauthors}{Shen, et al.}

\begin{abstract}
Since the introduction of the original BERT (i.e., BASE BERT), researchers have developed various customized BERT models with improved performance for specific domains and tasks by exploiting the benefits of {\em transfer learning}. Due to the nature of mathematical texts, which often use domain specific vocabulary along with equations and math symbols, we posit that the development of a new BERT model for mathematics would be useful for many mathematical downstream tasks. In this resource paper, we introduce our multi-institutional effort (i.e., two learning platforms and three academic institutions in the US) toward this need: {\m}, a model created by pre-training the BASE BERT model on a large mathematical corpus ranging from pre-kindergarten (pre-k), to high-school, to college graduate level mathematical content. In addition, we select three general NLP tasks that are often used in mathematics education: prediction of knowledge component, auto-grading open-ended Q\&A, and knowledge tracing, to demonstrate the superiority of {\m} over BASE BERT. Our experiments show that {\m} outperforms prior best methods by 1.2-22\% and BASE BERT by 2-8\% on these tasks. In addition, we build a mathematics specific vocabulary `mathVocab' to train with {\m}. We discover that {\m} pre-trained with `mathVocab' outperforms {\m} trained with the BASE BERT vocabulary (i.e., `origVocab'). {\m} is currently being adopted at the participated leaning platforms: Stride, Inc, a commercial educational resource provider, and ASSISTments.org, a free online educational platform. We release {\m} for public usage at: https://github.com/tbs17/MathBERT.

\end{abstract}



\maketitle

\pagestyle{plain} 
\section{Introduction}\label{intro}

The arrival of transformer-based language model, BERT~\cite{Devlin2019BERT:Understanding}, has revolutionized the NLP research and applications. One strength of BERT is its ability to adapt to new domain and/or task through pre-training by means of so-called ``transfer learning." By taking an advantage of this benefit, therefore, researchers have adapted BERT into diverse domains  (e.g., 
FinBERT \cite{Liu2020FinBERT:Mining}, ClinicalBERT \cite{HuangClinicalBert:Readmission}, BioBERT \cite{Lee2020DataMining}, SCIBERT \cite{Beltagy2019SCIBERT:Text}, E-BERT \cite{Zhang2020E-BERT:E-commerce}, LiBERT \cite{Guo2020DeText:BERT}) and tasks (e.g., \cite{Sun2019HowClassification}, \cite{Shen2021ClassifyingBERT}, \cite{Choi2021MelBERTTheories}, \cite{Liu2019RoBERTa:Approach}, \cite{Gururangan2020DontTasks}) with improved performances.

In the domain of mathematics, as mathematical text often use domain or context specific words, together with math equations and symbols, we posit that mathematics-customized BERT would help researchers and practitioners 
sort out the meaning of ambiguous language better by using surrounding text to establish ``math" context. Further, such an improved context-aware understanding of language could help develop and improve solutions for challenging NLP tasks in mathematics.

In mathematics education, for instance, there are several general tasks that currently cause researchers/educators headaches: (i) large-scale knowledge component (KC, a.k.a. skill) prediction (denoted as $T_{kc}$), (ii) open-ended question answer scoring (i.e., auto-grading) (denoted as $T_{ag}$), and (iii) knowledge tracing (KT) correctness prediction (denoted as $T_{kt}$). 
For instance, the struggle with $T_{kc}$ (e.g., predicting the right mathematical skill for a given text description) is partly attributed to its tediousness and labor-intensive work for teachers/tutors to label all knowledge components in texts where they need to organize mathematical problems, or descriptions of instructional videos, etc. 
The traditional way to address this challenge of $T_{kc}$ is to use machine learning to classify them via feature extraction \cite{Karlovcec2012KnowledgeSystemb,Pardos2017ImputingContext,Patikorn2019GeneralizabilityTexts}, which has produced decent results. 

However, open-ended essay or mathematical problem questions are becoming less popular in students' assignments due to the difficulty of developing universal automated support in assessing the response quality, causing educators to favor multiple choice questions when evaluating their students. 
According to Erikson et al. \cite{Erickson2020TheFormat}, from 2010 to 2020, less than 15\% of the assigned open response problems in ASSISTments \cite{Heffernan2014TheTeaching} were ever graded by teachers. However, in general, open-ended questions are known to be able to provide critical evaluation in testing students' true critical thinking and understanding. Therefore, it is still important to develop an effective solution toward $T_{kc}$. 

Similarly, {\em Knowledge Tracing}, a very important task in the education domain, is defined as the task of tracing students' knowledge state, which represents their mastery of educational content based on their past learning activities. Predicting students' next question correctness as a KT task is, for instance, well studied \cite{Corbetr1995KnowledgeKnowledge,Thai-Nghe2010RecommenderPerformance,Liu2019EKT:Prediction,Lee2019CreatingAssess,Pandey2019ATracing} but these solutions tend to rely on high-dimensional sequential data. The current solutions are still not able to capture the complex nature of students' learning activities over extended periods of time.

Addressing this lack of general BERT-based language model in mathematics education, therefore, in this work, we introduce our effort across  two learning platforms (i.e., ASSISTments and K12.com) and three academic institutions (i.e., Penn State, WPI, and U. Arkansas) in the US: {\bf {\m}}, a model created by pre-training the BASE BERT model on a large mathematical corpus ranging from pre-kindergarten (pre-k), to high-school, to college graduate level mathematical content.
In light of the recent successes from transfer learning models such as ELMo \cite{Peters2018DeepRepresentations}, ULMFiT \cite{Howard2018UniversalClassificationb} and BERT \cite{Devlin2019BERT:Understanding}, we propose to use a BERT-like model to improve the solutions of the aforementioned three tasks in one shot, as BERT has been proven to have outstanding performance in various NLP tasks.

However, directly applying BERT to mathematical tasks has limitations. 
First, the original BERT (i.e., BASE BERT) was trained mainly on general domain texts (e.g., general news articles and Wikipedia pages). As such, it is difficult to estimate the performance of a model trained on these texts on tasks using datasets that contain mathematical text. 
Second, the word distributions of general corpora is quite different from mathematical corpora (e.g., mathematical equations and symbols), which can often be a problem for mathematical task related models. 

Therefore, we hypothesize that a special BERT model needs to be trained on mathematical domain corpora to be effective in mathematics-related tasks. That is, we further {\em pre-train} the BASE BERT on  mathematical corpora to build {\m}. Then, we use the pre-trained weights from {\m} to {\em fine-tune} on the mathematical task-specific text dataset for classification. 

We make the following contributions in this work:
\begin{enumerate}
\item We build {\m} by pre-training the BASE BERT on mathematical domain texts ranging from pre-k to high-school to graduate level mathematical curriculum, books and paper abstracts. We publicly release {\m} as a community resource at: 
\begin{itemize}
    \item \verb|https://github.com/tbs17/MathBERT| 
for codes  on how to further-train and fine-tune, and 
\item 
\verb|https://huggingface.co/tbs17/MathBERT|
for PyTorch  version {\m} and tokenizer.
\item AWS S3 URLs \footnote{http://tracy-nlp-models.s3.amazonaws.com/mathbert-basevocab-uncased/\\ http://tracy-nlp-models.s3.amazonaws.com/mathbert-mathvocab-uncased/} for Tensorflow version {\m} and tokenizer.
\end{itemize} 

\item 
We build and release  a custom vocabulary \verb|mathVocab| to reflect the different nature of mathematical corpora (e.g., mathematical equations and symbols). We compare the performance of {\m} pre-trained with \verb|mathVocab| to {\m} pre-trained with the original BASE BERT vocabulary.

\item We evaluate the performance of {\m} for three general NLP tasks, $T_{kc}$, $T_{ag}$ and $T_{kt}$, and compare its performance to five baseline models.
Our experiments show that solutions of three tasks using {\m} outperforms those using BASE BERT by 2-8\%.

\item We sketch the use cases of {\m} currently being adopted at two major learning management systems: ASSISTments and K12.com by Stride.
\end{enumerate}

\section{Related Work}

The state-of-the-art language model BERT (Bidirectional Encoder Representations From Transformer) \cite{Devlin2019BERT:Understanding} is a pre-trained language representation model that was trained on 16 GB of unlabeled texts, including Books Corpus and Wikipedia, with a total of 3.3 billion words and a vocabulary size of 30,522. Its advantage over other pre-trained language models such as ELMo \cite{Peters2018DeepRepresentations} and ULMFiT \cite{Howard2018UniversalClassificationb} is its bidirectional structure by using the \textit{masked language model} (MLM) pre-training objective\cite{Devlin2019BERT:Understanding}. 
The MLM randomly masks 15\% of the tokens from the input to predict the original vocabulary id of the masked word based on its context from both directions \cite{Devlin2019BERT:Understanding}. The pre-trained model can be used directly to fine-tune on new data for NLP understanding and inference tasks or further pre-trained to get a new set of weights for transfer learning.

\begin{table}[t]
  \caption{Corpora Comparison for DAPT BERT Models}
  \label{tab:xbert}
  \resizebox{\columnwidth}{!}{
  \begin{tabular}{cccc}
    \toprule
    Domain &Name&\# Tokens&Corpora\\
    \midrule
   \multirow{2}{*}{ General NLP} & \multirow{2}{*}{Original BERT}&\multirow{2}{*}{3.3B}&News article,\\&&& Wikepedia \\
   
    \hline
    \multirow{2}{*}{Bio Medicine} &\multirow{2}{*}{BioBERT}&\multirow{2}{*}{18B}&PubMed,\\&&& PMC articles\\
    \hline
    \multirow{2}{*}{Clinical Medicine} &\multirow{2}{*}{ClinicalBERT}&\multirow{2}{*}{2M (notes)}&Hospital\\&&& Clinical Notes\\
    \hline
    \multirow{2}{*}{Science} & \multirow{2}{*}{SciBERT}&\multirow{2}{*}{3.2B}&Semantic\\&&& Scholar Papers\\
    \hline
    \multirow{2}{*}{Job}&\multirow{2}{*}{ LiBERT}&\multirow{2}{*}{685M}&LinkedIn search query\\&&& profile, job posts\\
    
    \hline
   E-commerce& E-BERT&233M (reviews)&Amazon Dataset\footnote{
https://nijianmo.github.io/amazon/index.html}\\

\hline
   \multirow{2}{*}{ Finance} & \multirow{2}{*}{FinBERT}&\multirow{2}{*}{12.7B}&Reuters \\&&&News stories\\
   
   \hline
    \multirow{2}{*}{Mathematics} & 
    {\bf {\m}} 
    &\multirow{2}{*}{ 100M}&Math curriculum and  books,\\& (This Work)&& Math arXiv paper abstract\\

  \bottomrule
\end{tabular}}
\end{table}
\begin{table}[t]
  \caption{Corpora Comparison for TAPT BERT Models. * indicates that the number is an estimation based on 150 tokens/sentence}
  \label{tab:tapt-data-cf}
  \resizebox{\columnwidth}{!}{
  \begin{tabular}{cccc}
   
  \hline
    Domain &Dataset&\# Tokens&
    Task\\
   
  \hline
   \multirow{2}{*}{ BioMed} & ChemProt~\cite{Gururangan2020DontTasks}&1.5M* &relation classification \\
   &RCT~\cite{Gururangan2020DontTasks}&12M*& abstract sent. roles \\
   
    \hline
     \multirow{2}{*}{Comp. Sci.} & ACL-ARC~\cite{Gururangan2020DontTasks}&291,150* &citation intent \\
   &SCIERC~\cite{Gururangan2020DontTasks}&697,200*& relation classification \\
    \hline
     \multirow{2}{*}{News} & HyperPartisan~\cite{Gururangan2020DontTasks}&96,750* & partisanship\\
   &AgNews~\cite{Sun2019HowClassification,Gururangan2020DontTasks}&5.6M& topic\\
    \hline
 \multirow{2}{*}{ Reviews} & Yelp~\cite{Sun2019HowClassification}&25M& review sentiment  \\
   &IMDB~\cite{Sun2019HowClassification,Gururangan2020DontTasks}&14.6M&review sentiment  \\
  
\hline
   \multirow{2}{*} { Linguistics} & VUA-20~\cite{Choi2021MelBERTTheories}&205,425 & metophor detection \\
   &VUA-Verb~\cite{Choi2021MelBERTTheories}&5,873&metophor detection\\
    \hline
   { Mathematics} & KC~\cite{Shen2021ClassifyingBERT}&589,549 &  skill code detection \\
 
   \hline

\end{tabular}}
\end{table}

\begin{figure*}[!ht]
  \centering   
  \includegraphics[width=0.75\linewidth]{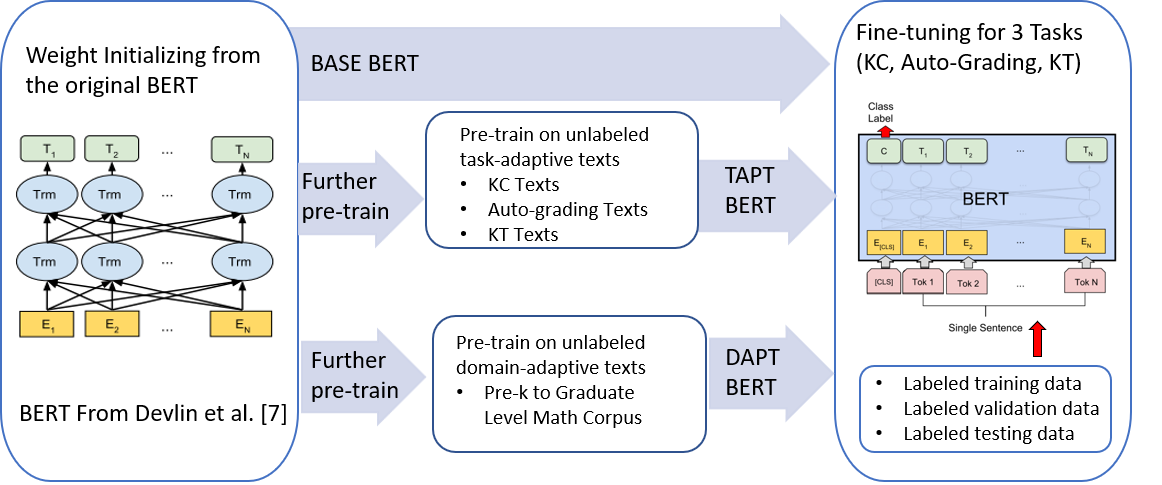}

  \caption{An illustration of training and fine-tuning process of BASE vs. TAPT vs. DAPT BERT models. The pre-training data are from this study. KC, Auto-grading, and KT Texts are task data for $T_{kc}$, $T_{ag}$, and $T_{kc}$ respectively.  }
  \label{fig:strategy}
\end{figure*}

The further pre-training process has become popular in the past two years as it is able to achieve better results than the fine-tuning only strategy. According to Gururangan et al. \cite{Gururangan2020DontTasks}, there are two styles of further pre-training on the BASE BERT \cite{Devlin2019BERT:Understanding}: (i) further pre-train the BASE BERT on a task-specific data set with tasks being text classification, question and answering inference, paraphrasing, etc. Gururangan et al. \cite{Gururangan2020DontTasks} call this kind of model a Task-adaptive Pre-trained ({\bf TAPT}) Model. (ii) further pre-train the BASE BERT on a domain-specific data set with domains being finance, bio-science, clinical fields, etc. Gururangan et al. \cite{Gururangan2020DontTasks} call this kind of model a Domain-adaptive Pre-trained ({\bf DAPT}) Model. Both TAPT and DAPT BERT models start the further pre-training process from the BASE BERT weights but pre-train on different types of corpora. TAPT BERT models pre-train 
on task-specific data, whereas DAPT BERT models pre-train on the domain-specific data before they are fine-tuned for use in any downstream tasks (see the process illustrated in Fig. \ref{fig:strategy}). 

The domain specific corpora that DAPT BERT models train on are usually huge (e.g. billions of news articles, clinical texts or PMC full-text and abstracts), which help DAPT BERT models achieve state-of-art (SOTA) performance in the corresponding domains. For example, FinBERT \cite{Liu2020FinBERT:Mining}, ClinicalBERT \cite{HuangClinicalBert:Readmission}, BioBERT \cite{Lee2020DataMining}, SCIBERT \cite{Beltagy2019SCIBERT:Text}. Other DAPT models such as E-BERT \cite{Zhang2020E-BERT:E-commerce} and LiBERT \cite{Guo2020DeText:BERT} 
not only further pre-trained on the domain specific corpora but also modified the transformer architecture to achieve better performance for the domain related tasks. 
A comparison between different domain-specific BERT models' corpora is shown in Table \ref{tab:xbert}. From the table, we can see that BioBERT was pre-trained on the largest set of tokens (18B) whereas our {\m} is pre-trained on the smallest set of tokens (100M). Although the scale of training data is much smaller than the BASE BERT, {\m} is still more effective in evaluating mathematics related tasks. 

There are also a few works that focus on TAPT models. Sun et al. \cite{Sun2019HowClassification} proposed a detailed process on how to further pre-train a TAPT BERT model and fine-tune it for three types of classification tasks (i.e., sentiment, question, and topic), achieving a new record accuracy. Shen et al. \cite{Shen2021ClassifyingBERT} pre-trained a TAPT BERT model to predict knowledge components and surpassed the BASE BERT accuracy by about 2\%. MelBERT \cite{Choi2021MelBERTTheories} further pre-trained the RoBERTa-base BERT on well-known public English data sets (e.g.,VUA-20, VUA-Verb) that have been released in metaphor detection tasks and obtained [0.6\%, 3\%] out-performance over the RoBERTa-base \cite{Liu2019RoBERTa:Approach}. Gururangan et al.\cite{Gururangan2020DontTasks} pre-trained RoBERTa-base \cite{Liu2019RoBERTa:Approach} on famous task data sets (e.g., Chemprot, RCT, ACL-ARC, SCIERC, Hyperpartisan, AgNews, and IMDB tasks) and obtained [0.5\%, 4\%] better performance than RoBERTa-base. Table \ref{tab:tapt-data-cf} presents the training data size for the aforementioned TAPT Models, showcasing
that TAPT models have much smaller training data size than the DAPT BERT models. In general, DAPT models usually achieve better performance (1-8\% higher) than TAPT models \cite{Gururangan2020DontTasks}. Although DAPT BERT models require more time and resource to train, they have wider applications than TAPT BERT models because they do not need to retrain in the case of different tasks, where TAPT BERT models tend to. 

In light of the aforementioned success, we also build a DAPT model, {\m}, that is further pre-trained from the BASE BERT model with a dedicated mathematical corpus. With the similar goal to our {\m}, we note that the work by \cite{PengMathBERT:Understanding} was also independently announced about the same time (i.e., \cite{PengMathBERT:Understanding} was submitted to arXiv while our {\m} was released to GitHub and Hugging Face, both in May 2021).
\cite{PengMathBERT:Understanding} also built a pre-trained BERT from the mathematical formula data and applied it on three formula-related tasks (i.e., math info retrieval, formula topic classification, formula headline generation). However, as they claimed, their BERT is the first pre-trained model for mathematical formula understanding and was only trained on 8.7 million tokens of formula latex data with the 400 surrounding characters from arXiv papers (graduate-level). Our {\m} is  pre-trained on 100 million tokens of more general purpose mathematical corpora including curriculum, books, and arXiv paper abstracts, covering all the grade bands from pre-k to college graduate-level. Our training data not only include formulas and their contexts but also more general mathematical instructional texts from books, curriculum, MOOC courses, etc. We consider our work has a potential to be widely used for ``general" mathematics-related tasks. For instance, {\m} in Hugging Face has been  downloaded more than 150 times since May 2021. As \cite{PengMathBERT:Understanding} has not released their code and model artifacts, we could not compare our results directly to theirs. We welcome further comparison and analysis by releasing all our code and model artifacts at \verb|https://github.com/tbs17/MathBERT|.

\section{Building {\m}} \label{training}

 \begin{figure*}[t]
  \centering  
  \subfloat[Content of a Math Book]{\includegraphics[width=0.55\linewidth]{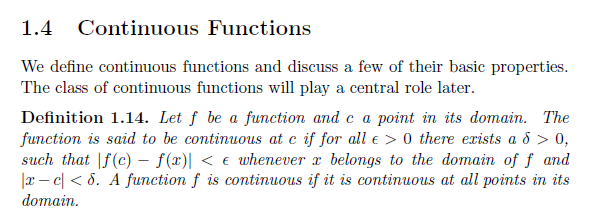}}\\
  \subfloat[Abstract of a Math arXiv Paper]{\includegraphics[width=0.7\linewidth]{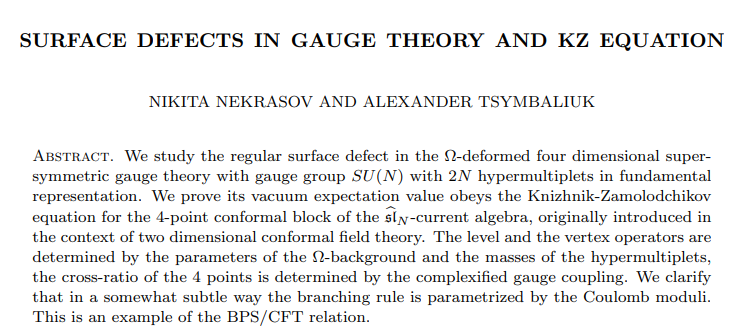}}\\
  \subfloat[Snippet of a Math Curriculum]{\includegraphics[width=0.7\linewidth]{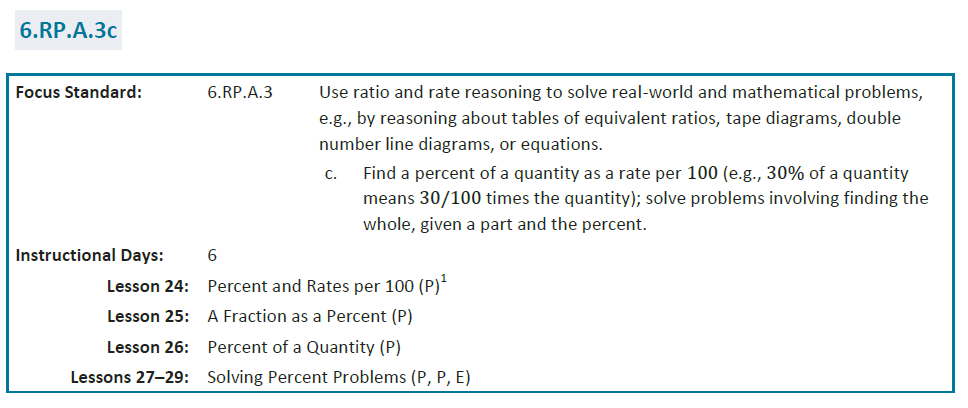}}\\

  \caption{Sample mathematical corpora text from math book, arXiv paper abstract, and curriculum
  }
  \label{fig:sample_text}
\end{figure*}

\subsection{Math Corpora} 
{\m} is pre-trained on mathematics related corpora that comprise mathematics curricula from pre-k to high school, mathematics textbooks written for high school and college students, mathematics course syllabi from Massive Online Open Courses (MOOC) as well as mathematics paper abstracts (see in Table ~\ref{tab:math_corpus}). We crawl these data from popular mathematics curriculum websites (illustrativemathematics.org, utahmiddleschoolmath.org, engageny.org), a free text book website (openculture.com), a MOOC platform (classcentral.com), and arXiv.org, with a total data size of around 3GB and 100 Million tokens. The mathematics corpora not only contain text but also mathematics symbols and equations. Among all these data, the text book data is in PDF format and we hence converted them into text format using the Python package \verb|pdfminer|\footnote{https://pypi.org/project/pdfminer/}, which preserves the mathematics symbols and equations (see sample text in Fig. \ref{fig:sample_text}).  
 
\begin{table}
  \caption{Math Corpus Details. Note all the corpus is in mathematics domain }
  \label{tab:math_corpus}
  \resizebox{\columnwidth}{!}{
  \begin{tabular}{cccl}
    \toprule
    Source &Math Corpora&Tokens\\
    \midrule
    arxiv.org & Paper abstract&64M \\
    classcentral.com & College MOOC syllabus &111K\\
    openculture.com & pre-k to College Textbook&11M\\
    engageny.org & Pre-k to HS Curriculum &18M\\
    illustrativemathematics.org &K-12 Curriculum&4M\\
    utahmiddleschoolmath.org & G6-8 Curriculum&2M\\
    ck12.org & K-12 Curriculum&910K\\
  \bottomrule
\end{tabular}}
\end{table}

\subsection{Training Details and Outcome}
To pre-train {\m} efficiently, we adopt a similar data processing strategy to the ROBERTa model, which threaded all the sentences together and split them into a maximum length of 512-token sequence sections \cite{Liu2019RoBERTa:Approach}. In other words, one sequence of data is longer than the original single sentence from the mathematics corpora.  Inspired by SciBERT \cite{Beltagy2019SCIBERT:Text}, we create a custom mathematical vocabulary (\verb|mathVocab|) using Hugging Face  \verb|BertWordPieceTokenizer|\footnote{https://huggingface.co/docs/tokenizers/python/latest/quicktour.html} with a size of 30,522 from the BASE BERT. We select 50 words from the same rank tier of \#2100 to \#2150 and discover that \verb|mathVocab| has more mathematical jargon than the original vocabulary (\verb|origVocab|) from BERT \cite{Devlin2019BERT:Understanding} (see in Table \ref{mathvocab}). 

\begin{table}[tb]
\caption{Vocabulary Comparison: origVocab vs. mathVocab. Tokens in blue are mathematics domain specific.}
\label{mathvocab}
\resizebox{\columnwidth}{!}{
\begin{tabular}{c|c}
\hline
Vocab Type& 50 Selected Tokens (from \#2100-\#2150)\\
\hline
\multirow{5}{*}{origVocab} & \#\#y, later, \#\#t, city, under, around, did, \\&such, being, used, state,  people, part,\\& know, against, your,  many, second,  university,\\& both,  national,\#\#er, these, don, known,  off,\\&  way, until, re, how,  even, get,\\& head, ..., didn, \#\#ly,  team, american,\\& because, de, \#\#l, born, united,\\& film, since,  still, long, work,  south, us \\
\hline
\multirow{5}{*}{mathVocab}&cod, exist, \#\#olds, \textcolor{blue}{coun}, \#\#lud, \#\#ments,\\&  \textcolor{blue}{squ}, \#\#ings, known, ele, \#\#ks, fe,\\&  minutes, continu, \textcolor{blue}{\#\#line}, \textcolor{blue}{addi},  small, \textcolor{blue}{\#\#ology},\\& triang, \#\#velop,  \#\#etry, \textcolor{blue}{log}, \textcolor{blue}{converg}, \\&\textcolor{blue}{asym}, \#\#ero, \textcolor{blue}{norm}, \#\#abl, \#\#ern,\\&  every, \#\#otic,  \#\#istic, \textcolor{blue}{cir}, \#\#gy, \\&  \textcolor{blue}{positive}, \textcolor{blue}{hyper}, dep,  \#\#raw, \#\#ange,  analy,\\&  \textcolor{blue}{equival}, \#\#ynam, call,  mon,  \textcolor{blue}{numerical},\\&  fam, \textcolor{blue}{conject},  large, ques, \#\#sible, \textcolor{blue}{surf}\\
\hline
\end{tabular}}
\end{table}

%


We use 8-core TPU machine from Google Colab Pro to pre-train the BASE BERT on the mathematics corpora. The largest batch size (bs) we can fit into the TPU memory is 128 and the best training learning rate (lr) is $5e-5$ with maximum sequence length (max-seq) of 512 for both {\m} with \verb|origVocab| and \verb|mathVocab|. We measure the effectiveness of training via Mask Language Modeling (MLM) accuracy (ACC), where the model predicts the vocabulary ID of the masked words in a sentence \cite{Devlin2019BERT:Understanding}. For training steps, we find both versions of {\m} reach their best result at 600K with MLM accuracy of above 99.8\% after a training time of 5 days each. We release {\m} model artifacts trained with \verb|origVocab| and \verb|mathVocab| in both Tensorflow and Pytorch versions (see in \verb|https://github.com/tbs17/MathBERT|). Specifically, one can use AWS S3 bucket URLs\footnote{http://tracy-nlp-models.s3.amazonaws.com/mathbert-basevocab-uncased\\ http://tracy-nlp-models.s3.amazonaws.com/mathbert-mathvocab-uncased} to download the Tensorflow version of model artifact. The Pytorch version can be downloaded from the Hugging Face Repo\footnote{\url{https://huggingface.co/tbs17/MathBERT}} 
or directly installed within the Hugging Face's framework under the name space ``\verb|tbs17|" using the code below. 
\begin{lstlisting}[language=Python]
from transformers import AutoTokenizer
from transformers import AutoModelForMaskedLM
# Download the MathBERT-basevocab
tokenizer = AutoTokenizer.from_pretrained("tbs17/MathBERT")
model = AutoModelForMaskedLM.from_pretrained("tbs17/MathBERT")
# Download the MathBERT-mathvocab
tokenizer = AutoTokenizer.from_pretrained("tbs17/MathBERT-custom")
model = AutoModelForMaskedLM.from_pretrained("tbs17/MathBERT-custom")
\end{lstlisting}

\section{Downstream Math NLP Tasks}

\subsection{Three Tasks}

We use three mathematical tasks mentioned in Section \ref{intro} to demonstrate the usefulness of {\m}. They can be formulated as follows:


\begin{itemize}
    \item KC Prediction ($T_{kc}$): a single sentence \textit{multinominal classification} problem (213 labels) with $I nput (I) \mapsto text$ and $Output (O) \mapsto KC$ (i.e., one of 213 labels).
    \item Auto-grading ($T_{ag}$): a two-sentence \textit{multinominal classification} problem (5 labels) with $I \mapsto Question\&Answer$ pair and $O \mapsto Score$.
    \item KT Correctness ($T_{kt}$): a two-sentence \textit{binary classification} problem with $I \mapsto Question\&Answer$ pair and $O \mapsto Correctness$.
\end{itemize}

\subsection{Task Data}
The three task data sets are noted as $D_{kc}$ for $T_{kc}$, $D_{ag}$ for $T_{ag}$, and $D_{kt}$ for $T_{kt}$, respectively. They are used not only to fine-tune for task classification but also for pre-training TAPT BERT models, which will serve as baseline models for {\m} in Section \ref{eval}.
All of  three data sets are provided from ASSISTments \cite{Heffernan2014TheTeaching}. We use the same mathematical problem data set as in  the best performing prior work \cite{Shen2021ClassifyingBERT} with 13,722 texts and 213 labels for KC prediction. The auto-grading task data is the same as in the best performing prior work \cite{Erickson2020TheFormat} with 141,186 texts to predict scores 1 to 5. The KT data is the text version (269,230 texts and 2 labels)  of the ASSISTments 2009 data\footnote{https://sites.google.com/site/assistmentsdata/home/assistment-2009-2010-data/skill-builder-data-2009-2010}, the numeric form of which was used by the best performing prior work \cite{Lee2019CreatingAssess}. 

Among the three data sets, $D_{kc}$ has the smallest number of records (13,722 rows) but the most unique labels (213 labels), whereas $D_{kt}$ has the largest number of records (269,230 rows) but the least unique labels (2 labels) (see in Table \ref{tab:task_data}). 
These three data sets were chosen due to their accessibility and we don't expect our results would be significantly better or worse if we choose other data sets. When fine-tuning, both the labels and texts are used (see Column 2 and 3) with split ratio of 72\% training, 8\% validating, and 20\% testing. When pre-training for TAPT BERT models, only the unlabeled texts are used for further pre-training without splitting (see Column 3).

Table \ref{task_text} provides examples from the three task data sets. In $D_{kc}$, the label `8.EE.A.1' represents a knowledge component (KC) code where `8' means Grade 8, `EE' is the skill name called `Expression and Equation', and `A.1' is the lesson code. There are total of 213 KC codes in $D_{kc}$ with each represented by a specific knowledge component. In $D_{ag}$, the label `5' is the grading score `5' for the answer in the text. There are total of 5 labels in $D_{ag}$ with `5' being the highest and `1' being the lowest. In $D_{kt}$, the label `1' means `correct' for the answer in the text. There are total 2 labels in $D_{kt}$ with another label `0' meaning 'incorrect' for student answers.

\begin{table}
  \caption{Task Data Details. KC: Knowledge Component, KT: Knowledge Tracing. All data from ASSISTments platform\cite{Heffernan2014TheTeaching}}
  \label{tab:task_data}
  \resizebox{\columnwidth}{!}{
  \begin{tabular}{cccccc}
    \toprule
    \multirow{2}{*}{Task} &
    \multirow{2}{*}{\#Labels} &
    \multirow{2}{*}{\#Texts} &
    \multicolumn{3}{c}{\#Fine-tune Split}\\\cline{4-6}
    &&&Train (72\%)&Validate (8\%)&Test (20\%)\\
    \midrule
    $D_{kc}$ &213&13,722&9,879&1,098&2,745 \\
    $D_{ag}$  &5&141,186&101,653&11,295&28,238\\
    $D_{kt}$ &2&269,230&193,845&21,539&53,846\\

  \bottomrule
\end{tabular}}
\end{table}

\begin{table}[tb]
\caption{Example texts of the three tasks with labels }
\label{task_text}
\resizebox{\columnwidth}{!}{
\begin{tabular}{cc
c}
\hline
Task Data &Label& Text\\
\hline
\multirow{4}{*}{$D_{kc}$}  & \multirow{4}{*}{8.EE.A.1}& Simplify the expression: (z2)2 \\&&Put parentheses around the power \\&& if next to coefficient, for example:\\&& 3x2=3($x^2$),x5=$x^5$ \\
\hline
\multirow{4}{*}{$D_{ag}$}& \multirow{4}{*}{5}& Q: Explain your answer \\&& on the box below.\\&&A: because it is the same shape,\\&& just larger, making it similar \\
\hline
\multirow{2}{*}{$D_{kt}$} &\multirow{2}{*}{1} &Q: What is 2.6 + (-10.9)?\\&&A: -8.3\\

\hline
\end{tabular}}

\end{table}

\subsection{Task Training and Fine-tuning}

We pre-train BASE BERT on the unlabeled texts of $D_{kc}$, $D_{ag}$, $D_{kt}$ to build TAPT BERT models and compare their performance to {\m}. The difference between TAPT and DAPT BERT training is illustrated in Fig. \ref{fig:strategy} where the input corpora is different. DAPT BERT models have much larger corpora whereas TAPT BERT models are more specific to tasks. We pre-train three TAPT models with \verb|origVocab| from the BASE BERT \cite{Devlin2019BERT:Understanding}. Among them, 
$TAPT_{kc}$ and $TAPT_{ag}$ reach the best results at 100K steps and $TAPT_{kt}$ reaches its best result at 120K steps with the MLM accuracy of above 99\%. Each of the TAPT models takes approximately 1 day to train. In addition to creating TAPT models pre-trained from BASE BERT, we also pre-train TAPT models from the {\m} weights, called {\m}+TAPT. They reach the best results at steps of 100K for both \verb|origVocab| and \verb|mathVocab| with the MLM accuracy of above 99.6\%. The {\m}+TAPT models also take approximately 1 day each to pre-train. We try to keep the MLM accuracy of TAPT Models similar to {\m} (see in Table \ref{tab:train_acc}). 
\begin{table}

  \caption{Training Steps and Accuracy: {\m} vs. TAPT vs. {\m}+TAPT}
  \label{tab:train_acc}
  \resizebox{\columnwidth}{!}{%
  \begin{tabular}{ccccc}
    \hline
    \multirow{2}{*}{Model}&
    \multirow{2}{*}{Task}& 
    \multirow{2}{*}{Steps}&
    \multicolumn{2}{c}{MLM ACC (\%)}\\
    \cline{4-5}
    &&&origVocab&mathVocab\\
    \hline
   
    {\m}&/ &600K&99.85&99.95 \\
    \hline
    \multirow{3}{*}{TAPT}
    &$T_{kc}$&100K &100&/ \\
    &$T_{ag}$&100K&99.10&/\\
    &$T_{kt}$&120K&99.04&/\\
    \hline
    \multirow{3}{*}{{\m}+TAPT}
    &$T_{kc}$&100K&100&99.99\\
    &$T_{ag}$&100K&99.95&99.96\\
    &$T_{kt}$&100K&99.67&99.68\\
    \hline

\end{tabular}}
\end{table}
 
For fine-tuning, we apply $D_{kc}$, $D_{ag}$, $D_{kt}$ onto BASE BERT, TAPT BERT, {\m}, and {\m}+TAPT models separately. Below is an example code for fine-tuning on task data set with {\m} weights and \verb|origVocab|.
\begin{lstlisting}[language=Python]
os.environ['TFHUB_CACHE_DIR'] = OUTPUT_DIR
python bert/run_classifier.py \
  --data_dir=$dataset \
  --bert_config_file=uncased_L-12_H-768_A-12_original/bert_config.json \
  --vocab_file=uncased_L-12_H-768_A-12_original/vocab.txt \
  --task_name=$TASK \
  --output_dir=$OUTPUT_DIR \
  --init_checkpoint=$MathBERT-orig_checkpoint \
  --do_lower_case=True \
  --do_train=True \
  --do_eval=True \
  --do_predict=True \
  --max_seq_length=512 \
  --warmup_step=200 \
  --learning_rate=5e-5 \
  --num_train_epochs=5 \
  --save_checkpoints_steps=5000 \
  --train_batch_size=64 \
  --eval_batch_size=32 \
  --predict_batch_size=16 \
  --tpu_name=$TPU_ADDRESS \
  --use_tpu=True
\end{lstlisting} 
We discover that hyper-parameter tuning has more to do with the task data instead of the model itself. In other words, the best hyper-parameter combinations are the same across {\m}, TAPT, and {\m}+TAPT but vary from task to task. Table \ref{tab:hyperpar} shows the optimal combinations of all the hyper-parameters for each task. This result is obtained after hyper-parameter search on lr $\in {\{1e-5, 2e-5, 5e-5, 8e-5, 1e-4\}}$, bs $\in {\{8, 16,32, 64, 128\}}$, max-seq $\in {\{128, 256, 512\}}$, and ep $\in {\{5, 10, 15, 25\}}$.

\begin{table}[t]
\caption{Optimal Hyper-parameter Combination for Task fine-tuning}
    \resizebox{\columnwidth}{!}{
    \label{tab:hyperpar}
    \begin{tabular}{ccccc}
    \hline
        Task & learning rate&batch size&max sequence length&epochs  \\
        \hline
        $T_{kc}$ &5e-5&64&512&25\\ 
        $T_{ag}$ &2e-5&64&512&5\\ 
        $T_{kt}$ &5e-5&128&512&5\\ 
        \hline
    \end{tabular}}
    \vspace{-\baselineskip}
    
\end{table}

\begin{table*}[tb]
  \caption{Performance Comparison: {\m} vs. Baseline Methods across Five Random Seeds. Bold font indicates best performance and underlined values are the second best. * indicates statistical significance. $\Delta$ shows relative improvement (\%) of {\m} over baselines. }

  \label{tab:perf}
  \begin{tabularx}{\textwidth}{YYYY|Y|YY}
    \hline
    \multirow{2}{*}{Method}&
    \multirow{2}{*}{Vocab}&
    \multicolumn{2}{c|}{$T_{kc}$ (\%)}&
    \multicolumn{1}{c|}{$T_{ag}$ (\%)}&
    \multicolumn{2}{c}{$T_{kt}$ (\%)}\\
    \cline{3-7}
    && F1&ACC&AUC&AUC&ACC\\
    \hline
   {Prior Best }
    &\multirow{2}{*}{/}&\multirow{2}{*}{88.69\cite{Shen2021ClassifyingBERT}}&\multirow{2}{*}{92.51\cite{Shen2021ClassifyingBERT}}&\multirow{2}{*}{85.00\cite{Erickson2020TheFormat}}&\multirow{2}{*}{81.82\cite{Lee2019CreatingAssess}}&\multirow{2}{*}{77.11\cite{Lee2019CreatingAssess}}\\
    (p)&&&&&\\
    
    \hline
    {BASE-BERT}
    &\multirow{2}{*}{orig}&\multirow{2}{*}{90.14}&\multirow{2}{*}{91.78}&\multirow{2}{*}{88.67}&\multirow{2}{*}{88.90}&\multirow{2}{*}{86.88}\\
     (b)&&&&&\\
    \hline
    {TAPT}
    &\multirow{2}{*}{orig}&\multirow{2}{*}{91.77}&\multirow{2}{*}{92.96}&\multirow{2}{*}{90.34}&\multirow{2}{*}{95.88}&\multirow{2}{*}{93.49}\\
    (t)&&&&&\\
    \hline
    {\m}
    &orig (o)&\textbf{92.67}&93.79&\underline{90.57}&96.04&94.07\\
    (m)&math (c)&92.51&93.60&90.45&95.95&94.01\\
    
    \hline
    {{\m}+TAPT }
    &orig (o) &92.54&\underline{93.82}&\textbf{90.73}&\underline{97.25}&\underline{95.52}\\
    (mt)&math (c) &\underline{92.65}&\textbf{93.92}&90.46&\textbf{97.57}&\textbf{95.67}\\
    \hline    
    \hline
    \multirow{2}{*}{$\Delta_{m-p}$ }
    &orig&+4.49\%&+1.38\%&+6.55\%&+17.38\%&+21.99\%\\
    &math&+4.31\%&+1.18\%&+6.41\%&+17.27\%&+21.92\%\\
    \hline
    \multirow{2}{*}{$\Delta_{m-b}$}
    &orig&+2.81\%***&+2.19\%***&+2.14\%***&+8.03\%***&+8.28\%***\\
    &math&+2.63\%***&+1.98\%***&+2.01\%***&+7.93\%**&+8.21\%***\\
    \hline
    \multirow{2}{*}{$\Delta_{m-t}$}
    &orig&+0.98\%***&+0.89\%***&+0.25\%**&+0.17\%&+0.62\%***\\
    &math&+0.81\%***&+0.69\%***&+0.12\%&+0.07\%&+0.56\%***\\
    \hline
    \multirow{2}{*}{$\Delta_{m-mt}$}
    &orig&+0.14\%&-0.03\%&-0.18\%&-1.26\%***&-1.54\%***\\
    &math&-0.15\%&-0.35\%&-0.01\%&-1.69\%***&-1.77\%***\\
    \hline
    $\Delta_{m^c-m^o}$&/&-0.17\%&-0.20\%&-0.13\%&-0.09\%&-0.06\%\\
    \hline
    $\Delta_{mt^c-mt^o}$&/&+0.12\%&+0.11\%&-0.30\%&+0.33\%**&+0.16\%\\
    \hline
\end{tabularx}
\end{table*}

\section{Evaluation of {\m}} \label{eval}

We denote {\m} pre-trained with \verb|origVocab| as {\m}-orig and {\m} pre-trained with \verb|mathVocab| as {\m}-custom. To evaluate their effectiveness across the tasks of $T_{kc}$, $T_{ag}$ and $T_{kt}$, we fine-tune {\m} on $D_{kc}$, $D_{ag}$ and $D_{kt}$ and compare the performance to the baseline models (see in Table \ref{tab:perf}). 
We group the baseline models into four categories: (1) Prior solutions with the best known performance, \cite{Shen2021ClassifyingBERT,Erickson2020TheFormat,Lee2019CreatingAssess}, (2) BASE BERT without any further pre-training,
(3) TAPT BERT models pre-trained on the task specific texts from BASE BERT weights, and (4) {\m}+TAPT models pre-trained on the task-specific texts from {\m} weights in both \verb|origVocab| and \verb|mathVocab| versions. 

We use both F1 and ACC (i.e., Accuracy) to measure $T_{kc}$ prediction results because traditionally, KC problems have been evaluated using ACC \cite{Karlovcec2012KnowledgeSystemb,Pardos2017ImputingContext,Patikorn2019GeneralizabilityTexts,Rose2005AutomaticAssessment}. We provide the additional measure (F1) to account for the imbalance in the KC labels in $D_{kc}$. In addition, we use Area-Under-the-Curve (AUC) to measure $T_{ag}$ because AUC is the typical measure used for the auto-grading problem. Finally, both AUC and ACC are used to measure $T_{kt}$ because historically both metrics were used for evaluation \cite{Lee2019CreatingAssess,Zhang2017DynamicTracing,Pandey2019ATracing,Piech2015DeepTracing}. 
After obtaining the best hyper-parameter tuning for each task from Table \ref{tab:hyperpar}, we run each model with five random seeds. We report the average value over five random seeds for each model and use t-tests to evaluate the significance of these results. A t-test is not applied to prior test results as we do not have the five random seeds results from the prior best method due to the lack of accessible codes.

In Table \ref{tab:hyperpar}, we note that {\m}-orig is about 1.38\% to 22.01\% better and {\m}-custom is about 1.18\% to 21.92\% better than the best prior methods across all metrics and tasks. In addition, {\m}-orig outperforms BASE BERT by about 2.14 \% to 8.28\%, all with statistical significance and {\m}-custom outperforms it by about 1.98\% to 8.21\% across metrics and tasks, all with statistical significance. Both versions of {\m} out-performs TAPT BERT models by [0.07\%,0.98\%] relatively with statistical significance for all tasks. We see both versions of {\m} under-perform the {\m}+TAPT models by 0.03 \% to 1.77\% across all the metrics except for F1 score on $T_{kc}$ from {\m}-orig. However, only the metrics for $T_{kt}$ have obtained significance. This is expected as {\m}+TAPT was further pre-trained by adapting it to the task-specific data on top of the {\m} weights. 

In addition, the best performance for each task is all from {\m} related models. For example, for $T_{kc}$, the best F1 performance is from {\m}-orig followed by the second best from {\m}+TAPT-custom whereas the best and second-best ACC are from both of the {\m}+TAPT versions (\verb origVocab \& \verb mathVocab). For $T_{ag}$, we find the best AUC is from {\m}+TAPT-orig followed by {\m}-orig. For $T_{kt}$, the best and second best AUC and ACC are from both versions of {\m}+TAPT with {\m}+TAPT-custom having higher performance.


\section{Use Cases}

In this section, we describe the ongoing activities to incorporate {\m} into two popular learning platforms.

\subsection{ASSISTments}

\begin{figure}[tb]
    \centering
    \includegraphics[width=0.95\linewidth]{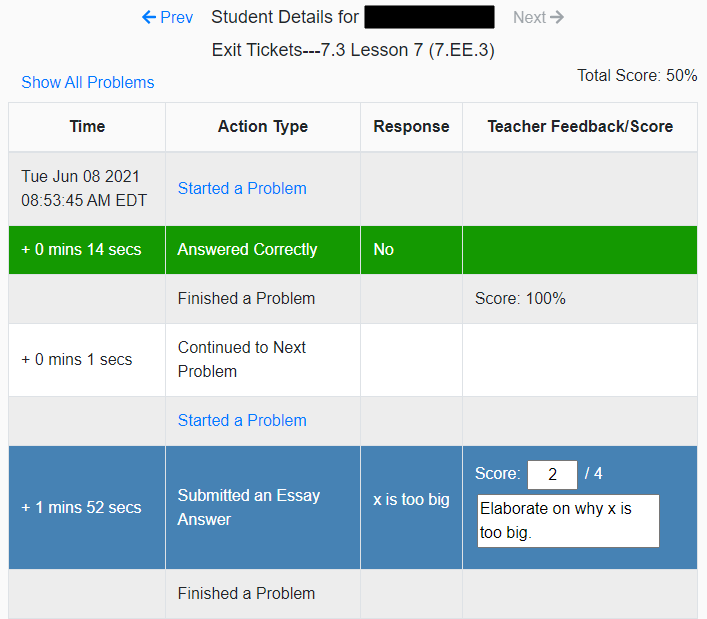}
    \caption{An open response in a student's report with the teacher's score and comment.}
    \label{fig:open_response}
\end{figure}

ASSISTments is an online learning platform that focuses on K-12 mathematics education. Within ASSISTments, teachers assign course work and view reports on their students. The reports show statistics on the class's performance and the responses of each student. Within the reports, teachers see a timeline of how each student progressed through the assignment and can grade students' open ended responses as well as leaving comments. Figure \ref{fig:open_response} shows an example of an open ended response within a student's report, together with the score and comment left by the teacher.

These open ended responses provide the first opportunity to use {\m} within ASSISTments. ASSISTments has recently begun using Sentence-BERT \cite{Reimers2019Sentence-BERT:BERT-Networks} to suggest grades to open response questions \cite{Baral2021ImprovingMathematics}. {\m} provides a more domain-specific BERT model for this task with high AUC. The similar task in our experiment $T_{ag}$ obtains 6.55\% higher in AUC than the prior best work \cite{Erickson2020TheFormat} which uses Sentence-BERT \cite{Baral2021ImprovingMathematics}, and can replace the current Sentence-BERT implementation. {\m} can not only provide teachers with suggested grades based on students' open ended responses, but also be used to suggest comments for teachers based on the content of the students' answers.

\begin{figure}[tb]
    \centering   
    \includegraphics[width=0.95\linewidth]{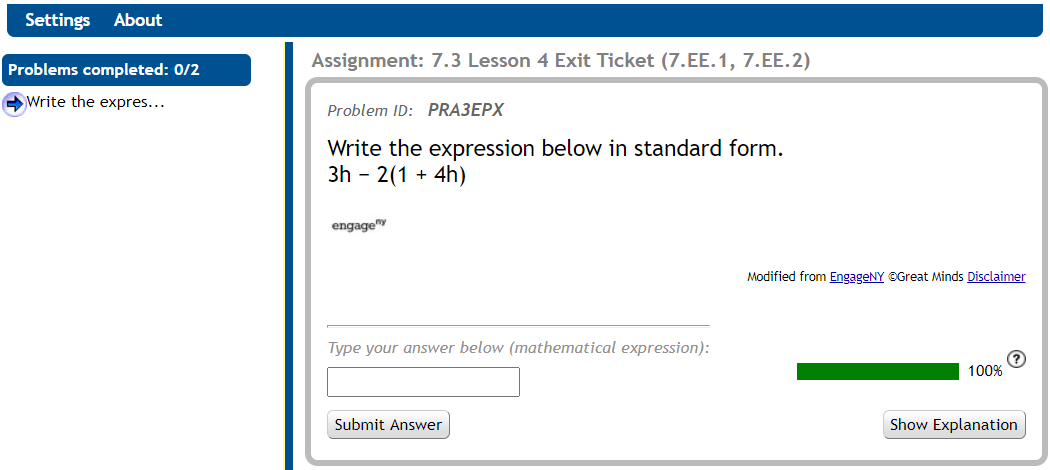}
    \caption{The ASSISTments Tutor, as seen by students when completing problem sets.}
    \label{fig:tutor}
\end{figure}

In addition to {\m}'s benefit to teachers using ASSISTments, {\m} can also be used to enhance the student experience. As students complete problem sets in the ASSISTments Tutor, shown in Figure \ref{fig:tutor}, they can be shown general educational material, such as YouTube videos, if they need additional guidance. {\m} can be used to identify relevant content by predicting the skills required to solve the problem. As the fine-tuning results for $T_{kc}$ using {\m}-orig shows, the F1 score and ACC for the top 3 predictions are 92.67\% and 93.79\% respectively. Relevant supplemental education material can then be selected and shown to the student. Identifying the skills required to solve a problem will also integrate well with ASSISTments' Automated Reassessment and Relearning System (ARRS) \cite{Wang2014TheMathematics}. This service automatically creates follow-up assignments for students when they fail to learn the material they were assigned. The purpose of the follow-up assignments is to test students' knowledge with problems similar to the ones the students previously got wrong. Although {\m} was tested on text prediction tasks such as $T_{kc}$, $T_{ag}$ and $T_{kt}$, it is not limited to only text prediction problems and can be applied to determine textual similarity, similar to the Semantic Textual Similarity Benchmark (STS-B) task from General Language Understand Evaluation (GLUE)\footnote{https://gluebenchmark.com/} which BASE BERT was evaluated on for its performance \cite{Devlin2019BERT:Understanding}. Therefore, we can use {\m} to automatically evaluate problems for similarity, either by determining the skills required to solve the problems, or by directly comparing problem texts.


\subsection{K12.com by Stride}


Stride, Inc that manages the learning platform of K12.com, 
is a leading education management organization that provides online education to American students from kindergarten to Grade 12 as well as adults. K-G12 math teachers rely on the Stride system to give math lessons, assign practice, home work, or exams, and grade them to provide feedback to students. Teachers have long been challenged by the time and effort they spend to grade and give feedback on open-ended math questions where various answers could be right and it is difficult to scale feedback for immediacy and volume.

\begin{figure}[t]
    \centering   
    \includegraphics[width=1.0\linewidth]{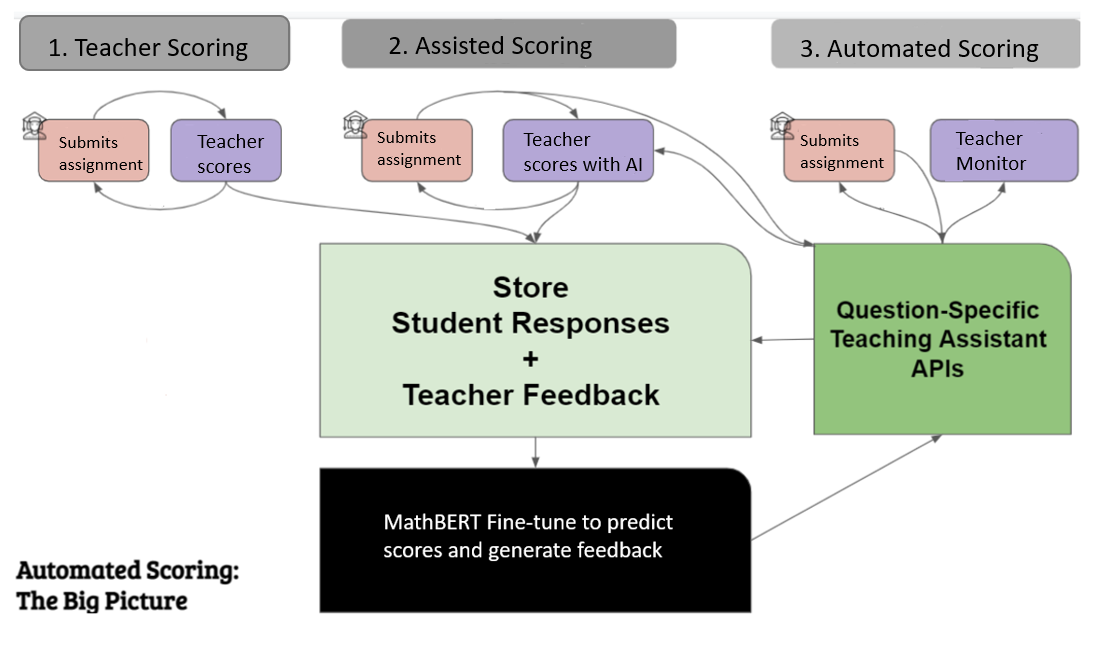}
    \caption{Stride auto-scoring pipeline using {\m}}
    \label{fig:auto-score}
\end{figure}

Therefore, Stride is considering an automatic scoring pipeline where they can train a model on their huge proprietary reservoir of open-ended responses and teacher feedback to automatically suggest scores and generate constructive feedback/comments for teachers to use. {\m} could be a nice fit for this model and play two roles: (i) {\m} fine-tunes on students' responses (input) with ground truth teacher scoring (label) to predict scores with high accuracy (as suggested by $T_{ag}$), and (ii) {\m} fine-tunes on the different scores (input) associated with teacher feedback (label) to predict/generate teacher feedback for a certain kind of score. For example, a student may only correctly answer part of the question and get a score of 3 out of 5, {\m} can recommend a feedback such as `You are very close! Can you tell us more?'. The prediction output from {\m} can then be wrapped into a question-specific teaching assistant API that prompts in front of students to guide them to reach the full score and truly master the knowledge component (see the pipeline in Fig. \ref{fig:auto-score}). 

The pipeline will be split into three phases: (i) collect data (i.e. responses, score, and feedback), (ii) use {\m} to fine-tune on the training data and predict scores and feedback, suggested to teachers via API. Teachers semi-auto grade and give feedback using {\m} suggested score and feedback. The final grade and feedback given to the students will then be sent back to the model to further fine-tine, and (iii) improve the accuracy of the question-specific teaching assistant API for fully automatic-scoring where teachers will only play a role in monitoring, reviewing the scores, and providing feedback.

\begin{figure}[h]
    \centering   
    \includegraphics[width=1.0\linewidth]{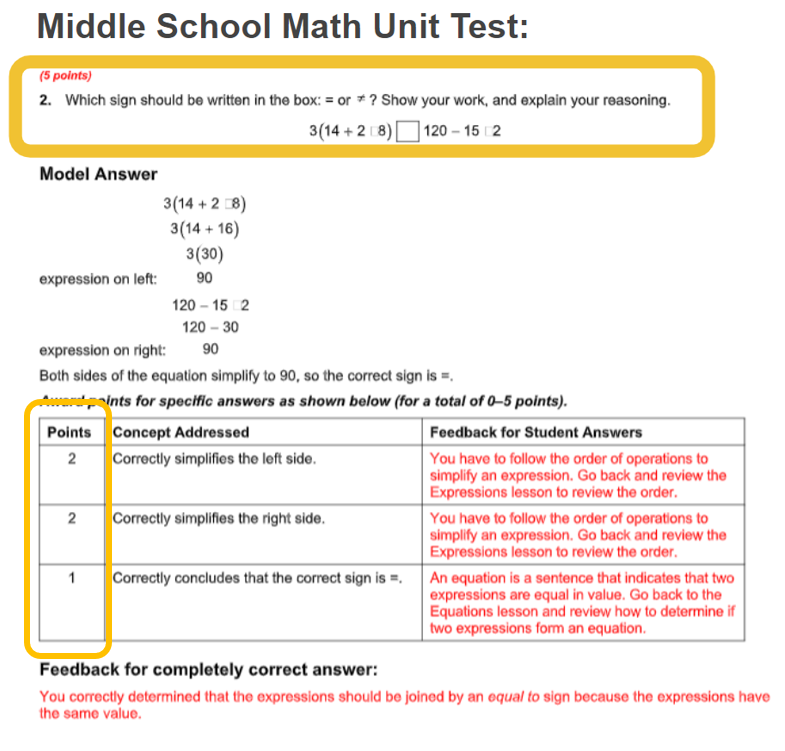}
    \caption{Stride auto-scoring model output in the unit test}
    \label{fig:interface}
\end{figure}

As a proof of concept, Fig.\ref{fig:interface} illustrates what {\m} will output after fine-tuning on the open-ended responses, scores, and feedback after phase 1. The red words are the feedback that the question-specific API will generate to guide students to achieve a full score. The points (in the yellow box) will be predicted by {\m} and automatically suggested to teachers. 

\section{Discussion and Limitation}

Although we have verified that {\m} is more effective than the BASE BERT for mathematics related tasks with a proportional improvement of [1.98\%, 8.28\%] with statistical significance, the effect from an in-domain vocabulary (\verb|mathVocab|) is not what we expect. As we see from Table \ref{tab:perf}, {\m}-custom has under-performed {\m}-orig when directly fine-tuned on, but outperformed {\m}-orig when further pre-trained on task specific data. However, t-tests show {\m}-orig is not significantly better than {\m}-custom and {\m}+TAPT-custom's out-performance over {\m}+TAPT-orig is only statistically significant for $T_{kc}$. 

As SciBERT \cite{Beltagy2019SCIBERT:Text} pointed out, the in-domain vocabulary is helpful but the out-performance over BASE BERT could be mainly from the domain corpus pre-training. Therefore, we argue that {\m} trained with \verb mathVocab   sometimes can be more beneficial than {\m} trained with \verb origVocab. In addition, we note that {\m} is not only applicable in text prediction tasks but also for other NLP understanding tasks such as paraphrasing, question and answering, and sentence entailment tasks. We evaluate {\m} for $T_{kc}$, $T_{ag}$, and $T_{kt}$ because  three tasks have been heavily studied and their test data are available to us.

In future, we plan to pre-train another {\m} on ``informal" mathematics-related texts as opposed to the formal mathematical content (e.g. math curriculum, book and paper) that the current {\m} is pre-trained on. We could potentially use such an informal {\m} to generate answers/conversations for mathematics tutoring chat bots.

\section{Conclusion}

In this work, we built and introduced {\m}-orig and {\m}-custom to effectively fine-tune on three mathematics-related tasks. Users can use the code from github to access the model artifacts. We showed that {\m} not only out-performed prior best methods by [1.18\%, 22.01\%], but also proportionally out-performed the BASE BERT by [1.98\%, 8.28\%] and TAPT BERT models by [0.25\%, 0.98\%] with statistical significance. {\m}-custom was pre-trained with the mathematical vocabulary (\verb|mathVocab|) to reflect the special nature of  mathematical corpora and sometimes showed better performance than {\m}-orig. {\m} currently is being adopted by two major learning management systems (i.e., ASSISTments and K12.com) to build automatic-scoring/commenting solutions to benefit teachers and students. 


\section{Acknowledgement}

The work was mainly supported by 
NSF awards (1940236, 1940076, 1940093). In addition, the work of Neil Heffernan was in part supported by NSF awards (1917808, 1931523, 1917713,  1903304, 1822830, 1759229), IES (R305A170137, R305A170243, R305A180401, R305A180401), EIR(U411B190024) and ONR (N00014-18-1-2768) and Schmidt Futures.


\bibliographystyle{ACM-Reference-Format}
\balance
\bibliography{references}





\end{document}